\documentclass{article}

\usepackage{templateArxiv}

\usepackage[utf8]{inputenc} 
\usepackage[T1]{fontenc}    
\usepackage{hyperref}       
\usepackage{url}            
\usepackage{booktabs}       
\usepackage{amsfonts}       
\usepackage{nicefrac}       
\usepackage{microtype}      
\usepackage{lipsum}
\usepackage{fancyhdr}       
\usepackage{graphicx}       
\usepackage[table,xcdraw]{xcolor}
\usepackage{times}

\usepackage{caption}
\usepackage{subcaption}
\usepackage{tikz}
\usetikzlibrary{arrows}
\usepackage{threeparttable}
\usepackage{multirow}
\usepackage{cancel}
\usepackage{tabularx}
\usepackage{paralist}
\usepackage{todonotes}

\usepackage{times}
\usepackage{graphicx}
\usepackage{caption}
\usepackage{subcaption}
\usepackage[numbers]{natbib}
\usepackage{algorithm}
\usepackage{algorithmic}
\usepackage{hyperref}

\usepackage{url}            
\usepackage{booktabs}       
\usepackage{amsfonts}       
\usepackage{nicefrac}       
\usepackage{microtype}      
\usepackage{color}
\usepackage{amsmath}
\usepackage{bm}
\usepackage{amsthm}
\usepackage{times}
\usepackage{amssymb}
\usepackage{xr}
\usepackage{lscape}
\usepackage{dsfont}
\usepackage{bbm}
\usepackage{authblk}

\allowdisplaybreaks
\graphicspath{{figures/}}     

\title{Trusting Language Models in Education
\thanks{\textit{\underline{Citation}}: 
\textbf{Authors. Title. Pages.... DOI:000000/11111.}} 
}

\author[1,2,$\dagger$]{Jogi Suda Neto}
\author[1]{Li Deng}
\author[1]{Thejaswi Raya}
\author[1]{Reza Shahbazi}
\author[1]{Nick Liu}
\author[1]{Adhitya Venkatesh}
\author[1]{Miral Shah}
\author[1]{Neeru Khosla}
\author[2]{Rodrigo Capobianco Guido}

\affil[1]{CK-12 Foundation}
\affil[2]{Instituto de Biociências, Letras e Ciências Exatas, Unesp - Univ Estadual Paulista São Paulo State University}
\affil[$\dagger$]{Corresponding author, contact: jogi.neto@ck12.org}
\begin{document}
\maketitle

\begin{abstract}
Language Models are being widely used in Education. Even though modern deep learning models achieve very good performance on question-answering tasks, sometimes they make errors. To avoid misleading students by showing wrong answers, it is important to calibrate the confidence - that is, the prediction probability - of these models. In our work, we propose to use an XGBoost on top of BERT to output the corrected probabilities, using features based on the attention mechanism. Our hypothesis is that the level of uncertainty contained in the flow of attention is related to the quality of the model's response itself.


\end{abstract}

\keywords{Confidence calibration \and Natural Language Processing \and Machine Reading Comprehension}

\section{INTRODUCTION AND BACKGROUND}


The innovation that Deep Learning has brought in the era of Big Data is considered a breakthrough, since those models gave practitioners the ability to solve a wide collection of difficult problems that Classical Machine Learning approaches couldn't perform well \citep{LeCun2015,Bahdanau2014Neural,Deng2013,DNN_SPMagazine12,Abdel-Hamid2014}. For example, we have seen great improvements in the medical area using computer vision \cite{vision_transformer,resnet}, and also in Natural Language Processing (NLP) \cite{BERT2018,sentence_transformers}, just to cite a few examples. This last area is going to be the focus of this paper. 

Specifically, at CK-12, we have a Question \& Answering (Q\&A) 
that starting with an input query, goes through several stages of processing. After the final stage we arrive at a set of candidate paragraphs that are likely to contain an answer to the query. This final stage is a softmax that ranks the paragraphs according to how likely they are to contain the correct answer.
we take the top-$3$ highest probabilities generated from the softmax output and show them to the user. 

The system is intended to receive all kinds of academic questions, and it should answer confidently when the question belongs to one of the domains the models were trained on, like biology, physics, math, etc. Some questions might be completely Out-of-Domain (see Fig. 1 (a), for some examples), or they might be slightly domain-shifted. In the latter case, we mean questions that have an intersection with the training topics, however a complete answer is not present at all in the corpus. For example, one may want to know about what is the Relativity Theory at a graduate level, and the model's predictions could be some very introductory answers at a high-school level of depth (Fig 1(b) illustrates some other examples to give a clear idea of domain-shifted questions). So, it's important to know when to abstain from answering a specific question, as this will mitigate the chances of misleading a student. In other words, a model's internal confidence should first be \textbf{reliable} for taking the decision to answer the question or not.

One major problem present in Deep Learning models is the confidence miscalibration. To be specific, let's consider a binary classification problem in the supervised learning setting. We know that, given an input instance represented by a finite feature-vector (in our case, it's a collection of tokens that are transformed into these vectors, also called word embeddings) of the form $x=(x_{1}, x_{2}, x_{3}, ..., x_{n}), n \in \mathbb{N}$, after a series of nonlinear transformations over each layer, generally the Neural Network outputs a value of a sigmoid activation (or softmax activation, in case of a multiclass classification problem), which can be interpreted as the internal confidence the model has of its prediction. For instance, given a sample of $n = 100$ queries, if the model outputs an internal score of $80\%$ for each of them, it's expected that about 80 of the queries will have a correct answer from the model. The problem is that DL models usually suffer from highly miscalibrated scores, so a high confidence but wrong prediction (or a low confidence but right prediction) might happen; see Fig. 1 for instance. Note that a model's empirical accuracy might be high, but the reliability of the internal probability scores may still be uncalibrated. In a wide variety of applications, especially ones involving high-risks such as fraud detection and self-driving cars, to cite a few - a miscalibrated prediction can have a really high cost and therefore is not tolerable. In other words, the challenge is knowing when the model is unlikely to have the right answer. By having a calibrated score, we can set a threshold to decide when to refuse answering the question. The problem of miscalibration in DL models is explored in further details in \cite{temp_scaling} and also \cite{li_calibration}.

The approach we propose here in this paper to overcome this problem is to train an XGBoost on top of the 
final softmax stage, receiving as input features related to the 
preceding encoder system, and the tokens. This is a similar approach to \cite{stanford}; our main improvement, however, comes from adding new attention-based features. First, we begin by interpreting the attention as a \textbf{flow}. Consider the attention from [CLS] to the [SEP] token, for instance; each layer can then be interpreted as a discrete-time step, where the attention is evolving as a flow. We consider that the attention flow is of important relevance as a feature for the calibrator (and confirm this in our experiments in section V), as it captures how the BERT-based model is relating the semantics (embeddings) of the answer to the question over time. As we see in section VI, our approach yields improved results over the previous work from \cite{stanford}.

\section{Related works}

\par Previous works in RC to improve confidence calibration have been done. In (\cite{temp_scaling}), several methods for confidence calibration of Deep Neural networks have been discussed. Their best method is the famous Temperature scaling, where the logits before the softmax activation are divided by some constant T, found via NLL optimization on the validation set; this has the effect of increasing the entropy of the model's output probability distribution. However, this doesn't change the AUC scores, as observed in section IV. For the sake of comparison, we also present the difference in results with respect to other approaches proposed in the paper, i.e.: Isotonic regression, Platt scaling and Temperature scaling.
\par In (\cite{towards_confident_mrc}), a Gradient Boosting Machine is proposed to calibrate the confidences. This GBM is trained on several features, including query and answers token length, softmax probability of start and end of answer spans, and even features based on the Query Embedding Gradients of the model, among others. The problem is that our RC model doesn't work by generating answer spans. Instead, given a query and a paragraph, it outputs the probability that the latter contains a response to the query. Another problem is the increase in complexity for calculating gradients, which is crucial for CK-12 and a lot of systems.
\par In (\cite{stanford}), the most similar to our approach, they use an XGBoost with multiple features: softmax scores, attention-based, and features related to the query and answer token lengths. Our improvements come from adding relevant attention features by interpreting them as a \textbf{flow}. From the attention flow, one can extract different informations, as presented in section IV. As a side note, in the paper, the authors put emphasis only on knowing when to refrain. We show, additionally, that not only the new calibrator has a better threshold for non-answerability, but the answerable questions also become better calibrated. This is shown in section V, in the reliability plots.
\par Now, we briefly explain the methods used in (\cite{temp_scaling}) that were also tested in this work.
\subsection{Platt Scaling}

\par (\cite{platt1999probabilistic}). The main idea here is to use a parametric approach for calibration. Basically, the prediction labels are used to train a logistic regression model. This calibrator model is trained on the \textbf{validation} set to return calibrated probabilities. For Deep Learning models, this approach has the following output, after learning parameters $a$ and $b\;\epsilon\;\mathcal{R}$ from optimizing NLL loss: 

$$\hat q_{i} = \sigma\left(az_{i} + b\right)$$

\par One important detail is that this is a post-processing calibration method, that is, the Neural Network's parameters are frozen during optimization of $a$ and $b$.

\subsection{Temperature Scaling}

\par Here, we have a specific case of Platt Scaling, since what this method does is learning a scalar parameter $T>0$ (also found via NLL optimization), and produces the following output:

$$\hat q_{i} = \max_{k}\sigma_{SM}\left(\frac{z_{i}}{T}\right)^{k}$$

\par Given $n$ logit vectors $z_{1},...,z_{n}$ and labels $y_{1},...,y_{n}$, then there is a unique $T$ that corresponds to the unique solution for the following entropy maximization problem:

\begin{equation}
\begin{aligned}
\begin{array}{rll}
\max_{T} & -\sum_{i=1}^{n} \sum_{k=1}^{K} T\left(\mathbf{z}_{i}\right)^{(k)} \log T\left(\mathbf{z}_{i}\right)^{(k)} \\
\text { subject to } & T\left(\mathbf{z}_{i}\right)^{(k)} \geq 0 \quad \forall i, k \\
& \sum_{k=1}^{K} T\left(\mathbf{z}_{i}\right)^{(k)}=1 \quad \forall i \\
& \sum_{i=1}^{n} z_{i}^{\left(y_{i}\right)}=\sum_{i=1}^{n} \sum_{k=1}^{K} z_{i}^{(k)} T\left(\mathbf{z}_{i}\right)^{(k)} .
\end{array}
\end{aligned}
\end{equation}

\par For which a detailed proof is found in (\cite{temp_scaling}). Essentially, it's preventing the logits before the activation to be pushed into extreme boundary regions of the softmax, and hence it's raising the entropy of the model's output. One caveat is that, since Temperature Scaling doesn't change the softmax maximum value (that is, it preserves the order between the outputs), it doesn't change the accuracy of the model, and hence also doesn't improve AUC, as we observe in Section VI.

\subsection{Isotonic Regression}

\par Unlike Platt Scaling, this is a nonparametric regression method for calibration (\cite{zadrozny2002transforming}), a piecewise constant function $f$ is learned to transform the uncalibrated outputs of a model: $\hat q_{i} = f(\hat p_{i})$. The optimization problem is formally written as:

\begin{equation}
\begin{aligned}
& \min_{\substack{M \\ \theta_{1}, ..., \theta_{M} \\ a_{1}, ..., a_{M+1}}} \sum_{m=1}^{M}\sum_{i=1}^{n}\boldsymbol{1}(a_{m} \leq \hat p_{i} < a_{m+1})(\theta_{m} - y_{i})^{2} \\
& \text{subject to }\;\; 0 = a_{1} \leq a_{2} \leq ... \leq a_{M+1} = 1 \\
& \;\;\;\;\;\;\;\;\;\;\;\;\;\;\;\;\; \theta_{1} \leq \theta_{2} \leq ... \leq \theta_{M}.
\end{aligned}
\end{equation}

\par Where $M$ represents the number of intervals, each $a_{1},...,a_{M+1}$ is an interval boundary, and $\theta_{1},...,\theta_{M}$ are the function values. Basically, $f$ is found such that it minimizes the square loss $\sum_{i=1}^{n}(f(\hat p_{i}) - \hat y_{i})^{2}$.

\section{Q\&A system}

The CK-12 Q\&A system is fundamentally based around variations of the BERT language model that have been optimized to perform well on the CK-12 corpus of academic content. The straightforward application of BERT in Q\&A often follows the instructions provided by its authors in the form of fine-tuning the vanilla BERT on the SQUAD data set. For a given question, the output of this system would be the span of text from the available corpus that likely constitutes an answer to the query. While this might be suitable for trivia type questions (e.g. ``What is the capital of Brazil?''), we found that for academic questions it leaves out important context which is a crucial part of the answer. For instance, in response to the question ``How many different types of volcano are there?'', a model fine-tuned on SQUAD might answer with ``Four kinds''. Which is indeed the correct answer, but incomplete. A better answer might look like ``Four kinds: cinder cones, composite volcanoes, shield volcanoes, and lava domes''. 

To that end, we devised our Q\&A system to answer with complete paragraphs so as to provide the students with enough context around the question. Starting with a query from the student, we employ multiple BERT language models to encode the question and try to find a matching paragraph for it. Specifically, for a given query $q$, the system outputs a conditional distribution over the available paragraphs, given $q$: $P(p_j|q)$ for $p_j$ the $j^{th}$ paragraph. From this distribution, we pick the maximum (or top $N$ maximum) likelihood paragraphs. Therefore, effectively the answer comes in the form of

\[\mbox{arg}\,\text{max}_{j}\,P(q_j|p)\]

The straightforward application of above will always pick an answer for any query, and fails to properly navigate the various situations where the question cannot be answered (e.g. an out of domain question). Therefore it is necessary to have additional measures in place to sort out a maximum likelihood candidate paragraph that truly answers the question, from one that does not.

\subsection{Vaswani's original formulation}

An important mechanism present today in the vast majority of the BERT-based models is the concept of \textbf{attention}. It is a mathematical formulation in the model's architecture that allows it to capture how itself is paying attention (and hence the name) to different tokens in the text. It is based on the idea that each word should have an importance to the meaning of the text. It can be described by mapping a query and a pair of key-value to some output, where all of these are vectors. The output is written as a weighted sum of the values, where the weights are given by a compatibility function between the query and the respective key. For Transformers, specifically, the model uses a similar concept called \textbf{Scaled Dot-Product Attention}, that was originally proposed by \cite{vaswani}, and its input consists of queries and keys of dimension $d_{k}$, and values of dimension $d_{v}$. A dot-product between $Q$ with all keys $K$ is computed; then, each result is divided by $\sqrt{d_{k}}$, and a softmax is applied to get the weights of the values.

In practice, a set of queries is assembled into a single matrix $Q$, and the same is done for the set of keys and values, represented in the end by matrices $K$ and $V$. Then, the output (which is also a matrix, then), is mathematically described as:
\begin{equation}
\begin{aligned}
Attention(Q,K,V) = softmax\left(\frac{QK^{T}}{\sqrt{d_{k}}}\right)V
\end{aligned}
\end{equation}
The authors argue that the reason for the scaling factor in the formula $\frac{1}{\sqrt{d_{k}}}$ is to prevent vanishing gradients, since the dot-products grow too large in magnitude as $d_{k}$ increases, thus pushing the softmax function to regions where its gradient becomes small.

Another important detail is that in BERT-based models, there are many attention functions running in parallel per each layer, thus projecting the queries, key and values to different learned linear projections with dimensions $d_{k}, d_{k}, d_{v}$. There is some evidence that each instance of this function - also called an attention head - contributes to understanding different parts of the semantic \cite{tenney2019bert,clark2019does}. In our model, specifically, we have 12 heads. See \ref{fig:Multi head attention} for a visual explanation.
\\
\\
\\
\begin{figure}[H]
    \centering
    \includegraphics[scale=0.5]{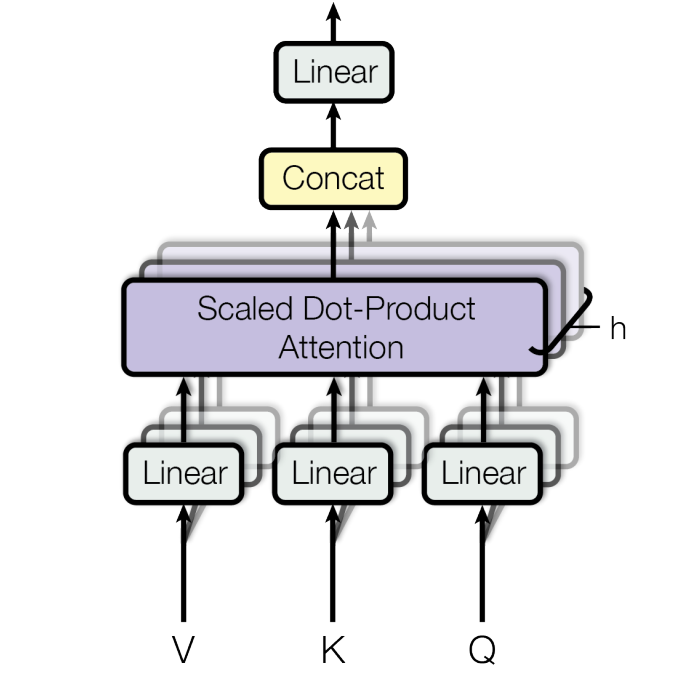}
    \caption{BERT's multihead attention mechanism illustrated. \cite{vaswani}}
    \label{fig:Multi head attention}
\end{figure}
~\\
\\

\section{Design of confidence measure and its features}

Given the needed background, in this work, our improvements come from interpreting the attention as a flow. Consider the attention from [CLS] to the [SEP] token in the first layer, for the first head, for instance; this gives us a scalar value. Each layer can then be interpreted as a discrete-time step, where the attention is evolving as a flow. We consider that the \textbf{attention flow} is of important relevance as a feature for the calibrator (and confirm this in our experiments in section V), as it captures how the BERT-based model is relating the semantics (embeddings) of the answer to the question over time.

\subsection{Shannon's Entropy}

The idea of entropy has been proposed in multiple fields through history, like in statistical mechanics \cite{jaynes1957information}, in Ergodic Theory \cite{kolmogorov1985new}, and also in Information Theory \cite{shannon1948mathematical}. Although all three definitions have connections, here we are interested in the latter. Given some random variable X and its distribution $p(x)$, Shannon's entropy \cite{shannon1948mathematical} is defined as:

$$H(p) = -\sum{p(x)\;log\;p(x)}$$

Shannon's entropy allows one to understand the notion of information as the unpredictability of $p$, or the number of bits needed to describe the distribution. So, $p$ in our case is the vector that represents the \textbf{attention flow}. Then, it is straightforward to see that measuring the entropy of $p$ is measuring the unpredictability, or the \textbf{information} (in the sense of Information Theory) contained in the \textbf{flow}. We hypothesize here that this should be a valuable feature for the calibrator, and we'll see in section VI that the experiments confirm it is among the most important features.  

\subsection{Delta scores}

Another way of incorporating the flow information ("information" now being used in its usual meaning) is by calculating the \textit{delta scores}. Given the vector of attention flow $A = \{A_{1}, A_{2}, A_{3}, ..., A_{N}\}$, being $N$ the number of layers, we define the delta of the flow as:

$$\delta_{A}=\{A_{i+1} - A_{i}:i=1, 2, 3, ..., N-1\}$$

We also tested the idea of delta scores for the model's top-3 probabilities, which ended up having a strong feature importance, as shown in section VI.

\subsection{Confidence measure}

\par As a way of comparing how well a model's internal probability is aligned with the true confidence measure, one useful metric that captures the notion of miscalibration is the \textbf{Average Calibration Error} (ACE), that works by partitioning the confidence and probability intervals (0-1) into $M$ equally-spaced bins, and then takes the average weighted sum of each bins' accuracy/confidence difference. Mathematically speaking, we have:

$$ACE = \sum_{m=1}^{M}\frac{|B_{m}|}{n}|acc(B_{m}) - conf(B_{m})|$$

\par Where $n$ is the number of samples. To emphasize the worst calibration error observed among the bins, we also measured the \textbf{Maximum Calibration Error} (MCE). Formally, we have:

$$MCE = \max_{m \epsilon \{1,2,...,M\}}|acc(B_{m}) - conf(B_{m})|$$

\par It's particularly useful in high-risk applications, where one wants the worst-case error to be as minimum as possible. Ideally, for a perfectly calibrated model, both ACE and MCE is 0.
\section{The New calibrator model}

Our proposed model is an XGBoost, which is an ensemble model, and for which its objective function is a logistic regression for binary classification; that is, it outputs a probability. The loss function is formally given as:
$$J(\theta)=\frac{1}{n}\left[\sum_{i=1}^{n}-y^{(i)} \log \left(p_{\theta}\left(x^{(i)}\right)\right)+\left(1-y^{(i)}\right) \log \left(1-p_{\theta}\left(x^{(i)}\right)\right)\right]$$
\par Where $n$ is the total number of samples, and $p_{\theta}(x^{i})$ is the probability the XGBoost model yields to the $i$-th data instance. In this work, we consider all the features used in \cite{stanford}, and also add the attention flow-based ones to get even better AUC and calibration. In total, we have the following features for the new calibrator:

\begin{itemize}
  \item Length of query and top-3 answer tokens.
  \item Top-3 softmax scores.
  \item Variance of top-3 softmax scores.
  \item Deltas of top-3 softmax scores.
  \item Attention-based feature from the base calibrator (see \cite{stanford} for more details).
  \item Entropy and deltas of attention flow.
\end{itemize}

\section{Experiment and results}

\subsection{Dataset}

For  training  our  calibrator,  we  considered  in  our  experiments  three  types  of  data:  in-domain,  domain-shifted  data,  and completely out-of-domain data. In practice, the Calibrator’s labeling receives 1 when there was at least one right answer among the top-3 most probable paragraphs, and 0 otherwise. By doing so, we have a simple binary classification problem, and yet we can identify when the model probably has at least one right prediction.

\subsection{Results}

Given the sufficient theoretical background, we present the results obtained at CK-12 real questions. First, we present the ACE, MCE and AUC scores for all the calibration methods tested:

\begin{table}[H]
\centering
\begin{tabular}{llll}
\hline
Method          & ACE       & MCE         & AUC  \\ \hline
CK-12 top-1     & 31.71\%   & 47.30\%       & 53\% \\
CK-12 top-2     & 34.22\%   & 50.20\%             & 60\% \\
CK-12 top-3     & 33.71\%   & 52.87\%             & 65\% \\ 
CK-12 top-1 (IR)    & 17.78\%     & 78.71\%           & 54\% \\
CK-12 top-2 (IR)    & 18.00\%     & 78.32\%           & 60\% \\
CK-12 top-3 (IR)    & 28.63\%     & 75.00\%           & 65\% \\
CK-12 top-1 (T = 2.67)     & 28.77\%  &  78.71\%            & 53\% \\ 
CK-12 top-2 (T = 2.67)     & 32.22\%  &  78.32\%            & 60\% \\ 
CK-12 top-3 (T = 2.67)     & 31.25\%  &  75.00\%            & 65\% \\
Platt scaling     & 11.11\%         &    16.97\%        & 70\% \\
Base calibrator & 8.45\%            &    14.41\%        & 74\% \\
New calibrator  & \textbf{3.99\%}   &    \textbf{8.09\%}          & \textbf{78\%} \\ \hline
\end{tabular}
\end{table}

\par From the table, we can see that the only methods that greatly diminish both ACE and MCE, while increasing AUC substantially, are the Platt scaling, the Base calibrator and the New calibrator. From those 3, the New calibrator outperforms all. Also, although Temperature scaling diminished ACE, it increased MCE, and didn't change AUC. We also present the reliability plots and the ROC curves below. 

\begin{figure}[H]
    \centering
    \includegraphics[height=300pt]{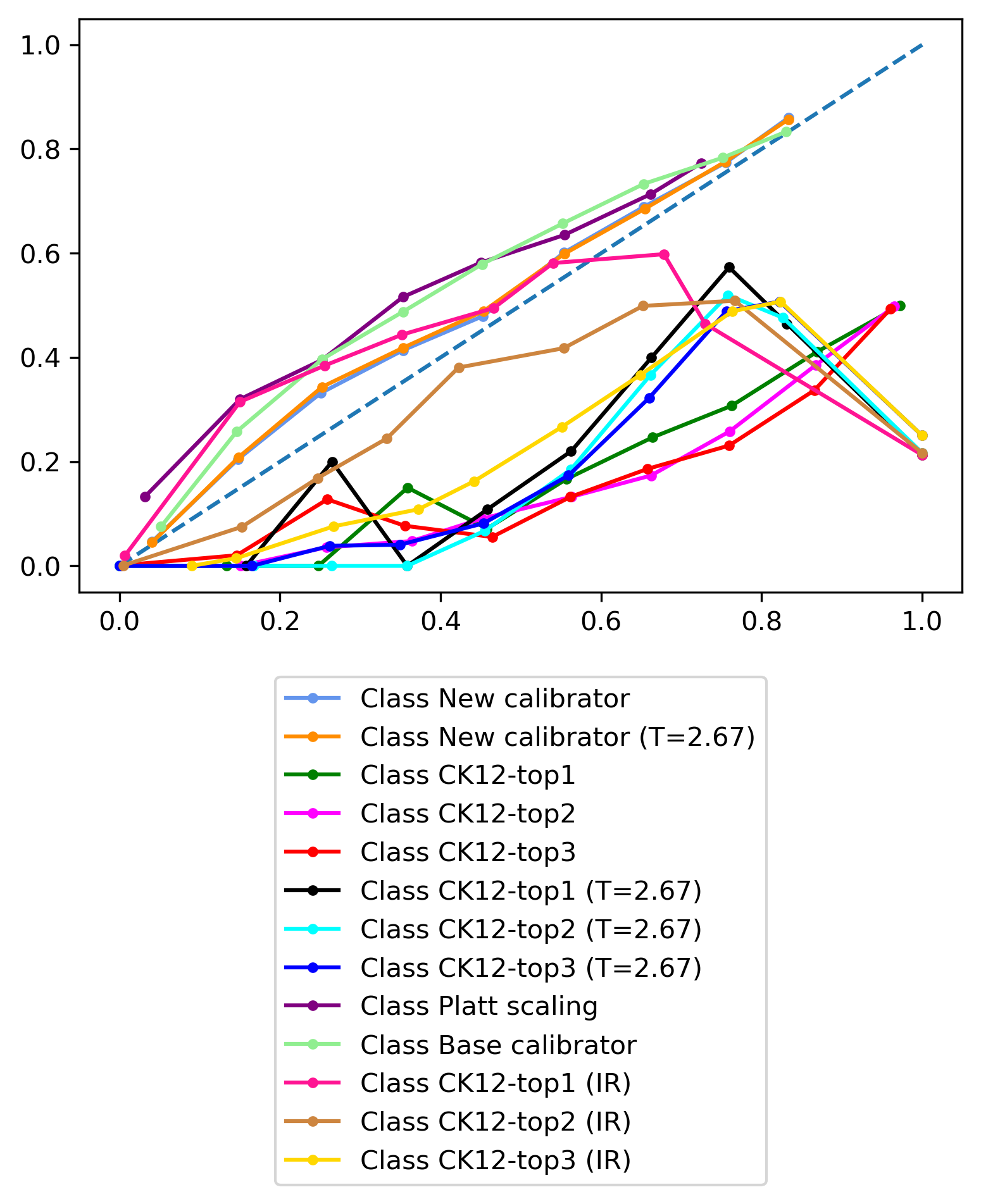}
    \caption{Reliability plots }
    \label{fig:Multi head attention}
\end{figure}

\begin{figure}[H]
    \centering
    \includegraphics[height=300pt]{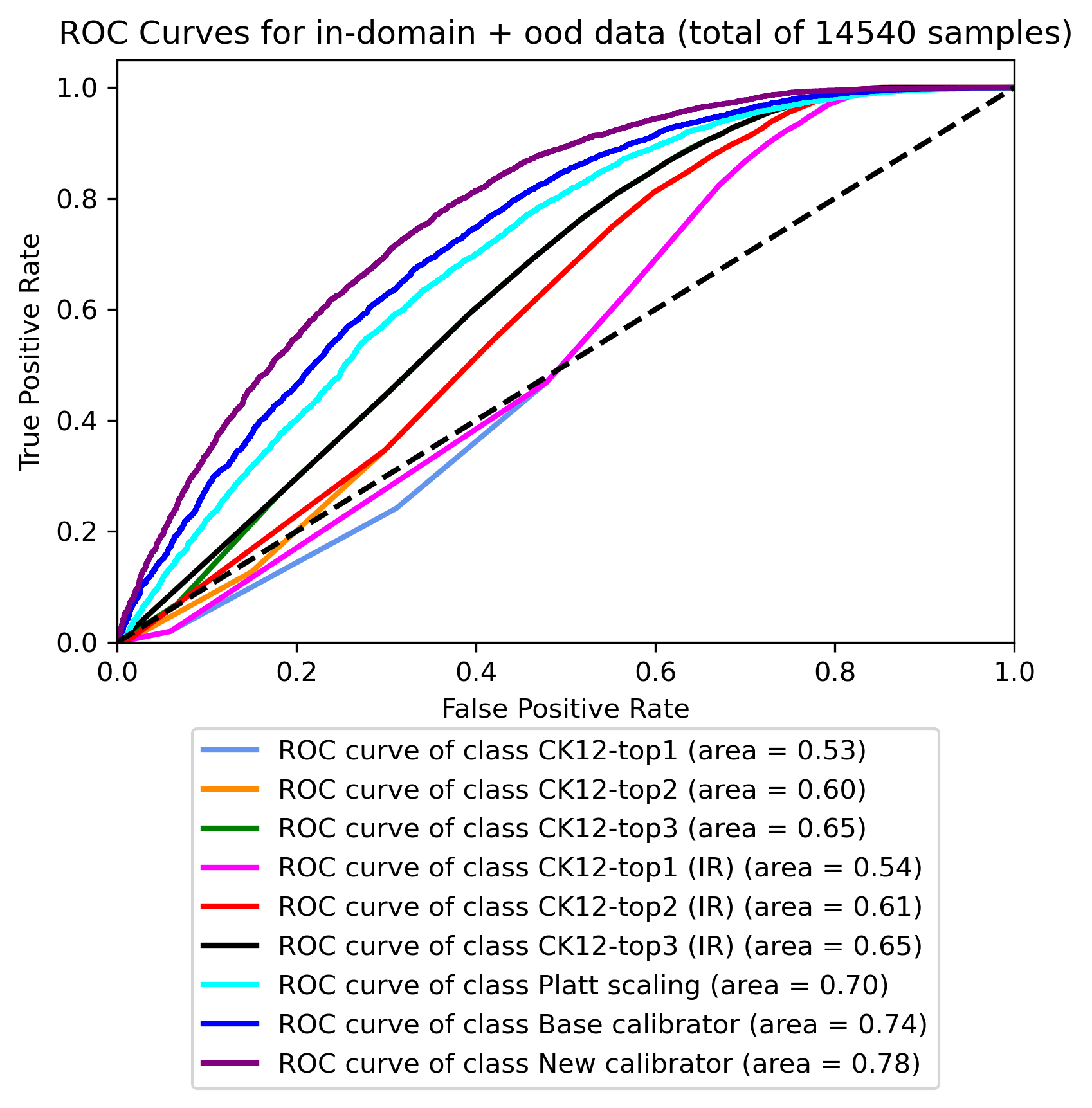}
    \caption{ROC curves with respectives AUCs}
    \label{fig:ROC}
\end{figure}

\par The top 5 feature importances from the XGBoost (New calibrator) are shown below:

\begin{table}[H]
\centering
\begin{tabular}{llll}
\hline
Feature          & Score        \\ \hline
CK-12 top-1 softmax score     & 52.0   \\
CK-12 top-2 softmax score     & 38.0  \\
CK-12 top-1 entropy of attention flow from [CLS] to [SEP]     & 32.0   \\ 
CK-12 top-3 softmax score    & 27.0     \\
CK-12 top-2 entropy of attention flow from [CLS] to [SEP]     & 22.0     \\ \hline
\end{tabular}
\end{table}
\par From this, we see that attention flow features and the raw softmax probabilities make the most important features for the calibrator. We also present the results for the top 20 most important features, for the sake of comparison: 
\begin{figure}[H]
    \centering
    \includegraphics[scale=0.65]{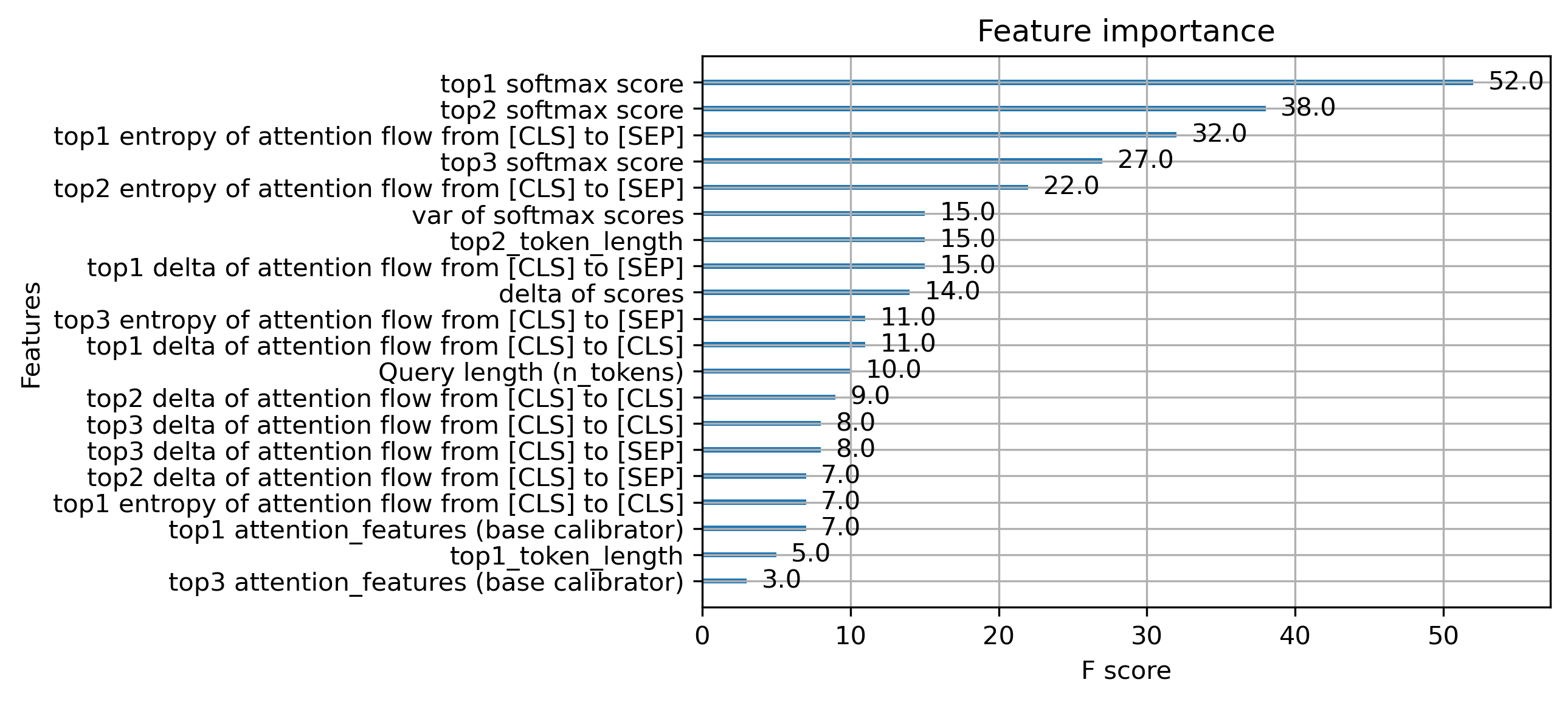}
    \caption{Feature importances for the New calibrator.}
    \label{fig:ROC}
\end{figure}

\subsection{Domain examples}

\par To illustrate the difference between in-domain (ID), domain-shifted (DS) and out-of-domain (OOD) questions, we present below some query examples, and the top 3 answers the model yields in each situation.

\begin{table}[H]
\centering
\resizebox{\textwidth}{!}{%
\begin{tabular}{|l|l|}
\hline
ID questions      & finetuned CK12 BERT predictions                                                                                                                                                                                                                                                                                                                                                                                                                                                                                                                                                                                                                                                                                         \\ \hline
Who was Albert Einstein? & \begin{tabular}[c]{@{}l@{}}1. could he be an aging rock star? he’s not a famous musician, \\ but he’s just as famous as many celebrities. his name is albert einstein, \\ and he’s arguably the most important scientist of the 20th century. \\ einstein really shook up science with his discoveries in the early 1900s. \\ that may sound like a long time ago, but in terms of the history of science, \\ it’s as though it was only yesterday.\\ \\2. albert einstein (1879–1955) deductive reasoning has helped us determine \\ that albert einstein is a mortal being.\\ \\3. einstein’s equation is possibly the best-known equation of all time.\end{tabular}                                                                                                                                                                                                                                                                                                                                                                                                                                                                                  \\ \hline
What is water made of?   & \begin{tabular}[c]{@{}l@{}}1. Water is a compound that consists of the elements hydrogen (h) and oxygen(o). \\     like other compounds, the smallest particles of water are called molecules. \\     each molecule of water (h2o) contains two atoms of hydrogen and one atom of\\     oxygen.\\ \\2. Water is probably one of the simplest compunds that you know. a water molecule\\     is made of two hydrogen atoms and one oxygen atom (figure below). all water\\     molecules have the same ratio: two hydrogen ions to one oxygen ion.\\ \\3. Water is a binary compound composed of hydrogen and oxygen. the hydrogen\\     and oxygen gases produced in the reaction are both diatomic molecules.\end{tabular} \\ \hline
\end{tabular}%
}
\end{table}

\begin{table}[H]
\centering
\resizebox{\textwidth}{!}{%
\begin{tabular}{|l|l|}
\hline
DS questions              & finetuned CK12 BERT predictions                                                                                                                                                                                                                                                                                                                                                                                                                                                                                                                                                                                                                                                                                                                                                                                                                                                                                                                                                                                                                                                                                                                                                                                                                                                                                                                                                                                                                                                                                                                                                                                                                                                             \\ \hline
How to solve a differential equation? & \begin{tabular}[c]{@{}l@{}}1. The equation relating to can be solved directly.\\ \\2. now substitute the know quantities into the equation and solve.\\ \\3. now substitute the known quantities into the equation and solve.\end{tabular}                                                                                                                                                                                                                                                                                                                                                                                                                                                                                                                                                                                                                                                                                                                                                                                                                                                                                                                                                                                                                                                                                                                                                                                                                                                                                                                                                                                                                                                      \\ \hline
What is plasma?                       & \begin{tabular}[c]{@{}l@{}}1. plasma is a state of matter that resembles a gas but has certain properties\\     that gases do not have. like a gas, plasma consists of particles of matter that \\     can pull apart and spread out, so it lacks a fixed volume and a fixed shape. \\     unlike a gas, plasma can conduct electricity and respond to a magnetic field. \\     that’s because plasma consists of electrically charged particles called ions, \\     instead of uncharged particles such as atoms or molecules.\\ \\ 2. plasma is a state of matter that resembles a gas but has certain properties that\\     gases do not have. like a gas, plasma consists of particles of matter than can pull \\     apart and spread out, so it lacks a fixed volume and a fixed shape. unlike a gas, \\     plasma can conduct electricity and respond to a magnetic field. that’s because plasma \\     consists of electrically charged particles called ions, instead of uncharged particles\\     such as atoms or molecules. this gives plasma other interesting properties as well. \\     for example, plasma glows with colored light when electricity passes through it.\\ \\3. plasma is a state of matter that lacks a fixed volume and a fixed shape and consists \\     of charged particles called ions. because it consists of charged particles, plasma can \\     conduct electricity and respond to a magnetic field.the sun and other stars consist of plasma. \\     plasma is also found naturally in lightning and the northern and southern lights. human-made\\     plasma is found in fluorescent lights, plasma tv screens, and plasma spheres.\end{tabular} \\ \hline
\end{tabular}%
}
\end{table}

\begin{table}[H]
\centering
\resizebox{\textwidth}{!}{%
\begin{tabular}{|l|l|}
\hline
OOD questions              & finetuned CK12 BERT predictions                                                                                                                                                                                                                                                                                                                                                                                                                                                                                                                                                                                                                                                                                                                                                                                                                                                                                                                                                                                                                                                                                                                                                                                                                                                                                                                                                                                                                                                                                                                                                                                                                                                             \\ \hline
Is Cristiano Ronaldo a better player than Neymar ? & \begin{tabular}[c]{@{}l@{}}1. could he be an aging rock star? he’s not a famous musician, \\ but he’s just as famous as many celebrities. his name is albert einstein, \\ and he’s arguably the most important scientist of the 20th century. \\ einstein really shook up science with his discoveries in the early 1900s. \\ that may sound like a long time ago, but in terms of the history of science, \\ it’s as though it was only yesterday.\\ \\2. young vs. old. does it matter?\\ \\3. the portuguese man o' war can deliver nasty stings with its tentacles.\end{tabular}                                                                                                                                                                                                                                                                                                                                                                                                                                                                                                                                                                                                                                                                                                                                                                                                                                                                                                                                                                                                                                                                                                                                                                                                                                                                                                                                                                                                                                                      \\ \hline
When is your birthday ?                       & \begin{tabular}[c]{@{}l@{}}1. what snack do you have after school?\\ \\2. these teens are good friends. like most teens, they spend more time with \\ one another than they do with family members. what do you \\ enjoy doing with your friends?\\ \\ 3. how old are you on earth? how old would you be if you lived on jupiter? \\ how many days is it until your birthday on earth? \\ how many days until your birthday if you lived on saturn?\end{tabular} \\ \hline
\end{tabular}%
}
\end{table}

\par From this, we can see different regimes. In the first situation, the user prompts a question to which the right answer is present in the corpus of possible responses. We call these queries in-domain questions. In another regime, the user might ask a question that contains only a partially right or related answer in the corpus. For example, the user might want to know about the General Theory of Relativity, and the corpus might contain an introductory high-school level of answer to this, but a complete dissertation is not present in the corpus at all. In the last situation, we consider completely out-of-domain questions, where no right answer at all is present in the corpus. These are questions like "Are you a bot?", or "When is your birthday?".

\section{Conclusion}
In this paper, we saw the importance of confidence calibration for Education Q\&A systems. By having a calibrated confidence, we mitigate the chances of misleading a student, because it allows one to not only have a precise estimation of how good an answer is, but also when the model should refrain from answering a question; these are crucial components in Education Q\&A systems. By adding the attention-flow-based features, we saw good improvements in our models, both in AUC by increase in 4 points, and also in ACE/MCE by reduction in 4.46\% and 6.32\%, respectively, all compared to the best previous approach in \cite{stanford}.

\section*{Declarations}
\subsection{Ethical Approval}
\par This work was supported by CK-12 Foundation.
\subsection{Conflict of interest}
\par The authors have no conflicts of interest to declare that are relevant to the content of this paper.
\subsection{Data availability}
\par Data availability is not applicable for this article.
\section*{Acknowledgements}
\par R.C.Guido gratefully acknowledges the grants provided by the Brazilian agencies ‘‘National Council for Scientific and Technological Development (CNPq)’’, Brazil and ‘‘The State of São Paulo Research Foundation (FAPESP)’’, Brazil, respectively through the processes 306808/2018-8 and 2021/12407-4, in support of this research.

\bibliographystyle{unsrt}  
\bibliography{references}  

\end{document}